\documentclass{article}
\usepackage{ijcai26}

\usepackage{times}
\usepackage{soul}
\usepackage{url}
\usepackage[hidelinks]{hyperref}
\usepackage[utf8]{inputenc}
\usepackage[small]{caption}
\usepackage{graphicx}
\usepackage{amsmath}
\usepackage{amsthm}
\usepackage{amssymb}
\usepackage{booktabs}
\usepackage{algorithm}
\usepackage{algorithmic}

\title{
LLM-assisted Semantic Option Discovery for Facilitating Adaptive Deep Reinforcement Learning 

}

\author{
}

\author{
Chang Yao$^1$
\and
Jinghui Qin$^2$
\and
Kebing Jin$^3$
\and
Hankz Hankui Zhuo$^4$\\  
}

\begin{document}

\maketitle

\begin{abstract}
Despite achieving remarkable success in complex tasks, Deep Reinforcement Learning (DRL) is still suffering from critical issues in practical applications, such as low data efficiency, lack of interpretability, and limited cross-environment transferability. However, the learned policy generating actions based on states are sensitive to the environmental changes, struggling to guarantee behavioral safety and compliance. Recent research shows that integrating Large Language Models (LLMs) with symbolic planning is promising in addressing these challenges. Inspired by this, we introduce a novel LLM-driven closed-loop framework, which enables semantic-driven skill reuse and real-time constraint monitoring by mapping natural language instructions into executable rules and semantically annotating automatically created options. The proposed approach utilizes the general knowledge of LLMs to facilitate exploration efficiency and adapt to transferable options for similar environments, and provides inherent interpretability through semantic annotations. To validate the effectiveness of this framework, we conduct experiments on two domains, Office World and Montezuma's Revenge, respectively. The results demonstrate superior performance in data efficiency, constraint compliance, and cross-task transferability.
\end{abstract}

\section{Introduction}

In recent years, Deep Reinforcement Learning (DRL) has achieved remarkable success in complex tasks such as gaming, robotic control \cite{andrychowicz2020learning}, and autonomous driving \cite{mnih2015human}. 
Moreover, the scope of DRL has expanded to encompass critical scientific and industrial domains, contributing to the emerging field of AI for Science and Technology.
Notable applications include optimizing chip floorplanning to accelerate hardware design \cite{mirhoseini2021graph}, discovering faster matrix multiplication algorithms for computational efficiency \cite{fawzi2022discovering}, and controlling tokamak plasmas in nuclear fusion research \cite{degrave2022magnetic}.
However, DRL still faces three major challenges in practical applications: low data efficiency, lack of interpretability in policies, and limited cross-environment transferability \cite{aradi2020survey}. Especially in structured environments with sparse rewards, delayed feedback, and environmental changes, traditional DRL methods often require massive interaction data and struggle to guarantee behavioral safety, interpretability, and transferability, which limits their deployments in real-world scenarios.
The research community has begun to explore hybrid architectures that integrate symbolic planning with reinforcement learning \cite{illanes2020symbolic,jin2022creativity}. Such methods introduce high-level symbolic action models and options, decomposing tasks into hierarchical sub-goals. This alleviates the sparse-reward problem, improves sampling efficiency, and provides a certain degree of policy interpretability through symbolic representations. 

Nevertheless, existing approaches still struggle with formalizing general knowledge according to learned policy and environmental adaptability when facing real-world complexity, due to the cross-task problems and misunderstanding of human intuition. \emph{Consider an office agent that has mastered navigation policies for a ``delivering coffee'' task. When switched to a new assignment not mentioned in the original environment, such as ``delivering juice'', an ideal process is that an agent transfers the acquired navigation policy to this new context without retraining.}  
Existing methods either treat the new task as a blank slate and relearn basic actions or utilize transfer learning to adapt to the new task. 

\begin{figure}[htbp]
   \centering
    \includegraphics[width=0.5\textwidth]{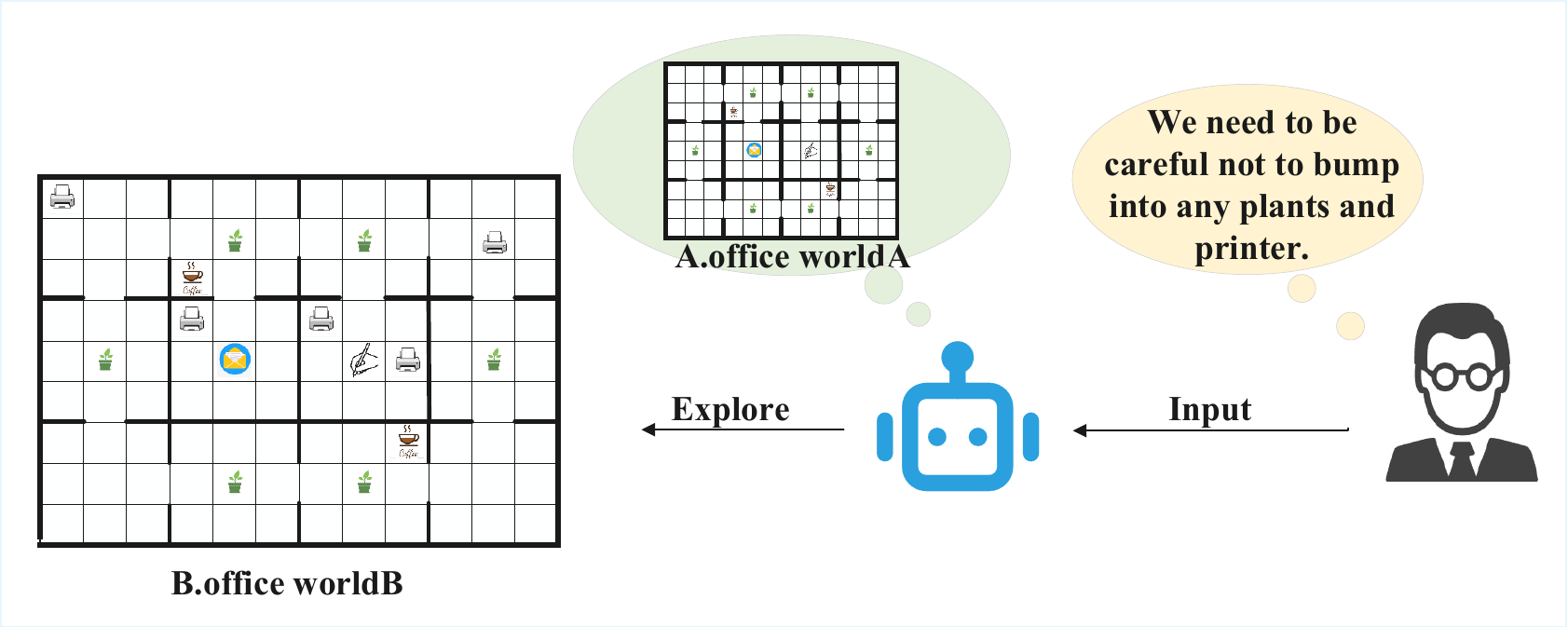}  
    \caption{Exploration of Natural Language Limitations Graph}
   \label{fig:Natural-Language}
\end{figure}

On the other hand, 
human prefer to utilize unstructured expression, such as natural language and images, to communicate with real-world environments. 
\emph{For example, as illustrated in the Figure\ref{fig:Natural-Language}, this example demonstrates the natural language requirements for an agent to explore a scenario, similar to the one in which the agent has abundant knowledge. The natural language instructions describe the differences between those scenarios. In the first scenario (Office World A), the agent learns the policy to take the coffee to targets, avoiding collisions with plants. And the second scenario is similar to the first scenario, but also includes printers, which require agents to avoid. Some DRL methods need to re-explore the second scenario to learn the policy due to the trivial change of environments. Moreover, in real-world human interactive applications, the kind of change can be done by semantic descriptions, upon the introduction of a printer as a new environmental variable, the user's command can be ``We need to be careful not to bump into any plants and printer.'' 
} 
Yet, traditional methods rely on rigid, expert-defined symbolic rules or pre-defined environments, lacking the ability to map such unstructured linguistic expressions into executable instructions.

Simultaneously, the emergence of Large Language Models (LLMs) offers a breakthrough opportunity. Pre-trained on massive datasets, recent studies have leveraged LLMs to enhance exploration efficiency by generating intrinsic rewards or proposing high-level sub-goals \cite{du2023guiding,tam2022semantic,xi2025rise}. However, those methods mostly utilize the computational and generative ability of LLMs to facilitate exploration\cite{shinn2023reflexion}, instead of finding the implied rules and formalizing general and reusable actions. In this paper, we want to explore whether the intrinsic world knowledge can serve as a valuable experience to assist agents in mitigating the blindness of learning from scratch, by utilizing the world knowledge and semantic information to translate natural language instructions into structured constraints and achieve policy adaptation to the new task.   


Therefore, in this paper, we propose LLM-SOARL (LLM-assisted Semantic Option-based Adaptive Reinforcement Learning), a closed-loop framework integrating LLMs, symbolic planning, and reinforcement learning. Our framework aims to provide an efficient and interpretable decision-making platform for scenarios with sparse rewards and high-level semantic interaction. The core contributions include:
\begin{itemize}

    \item We implement a LLM-driven semantic skill generation module that enables the agent to mine the implied general skills across different tasks.  
    \item We introduce a Constraint Adaptation module to help understand the scenario difference and compute adaptive policies according to intuitive linguistic instructions. 
    \item Leveraging the vast general knowledge embedded in LLMs, we utilize the intrinsic information as auxiliary knowledge to interpret environmental and option semantics, thereby facilitating the exploration process.
\end{itemize}

The framework aims to provide an efficient, safe, interpretable, and transferable decision-learning platform for task scenarios with complex constraints, sparse rewards, and high-level semantic interaction. We validate the superior performance of  LLM-SOARL in terms of data efficiency, constraint compliance, and cross-task transfer in typical environments such as Office World and Montezuma's Revenge.

\section{Problem Definition}
We define the reinforcement learning environment by a tuple \((C, I, G, P, A, S, F, \tilde{S}, \tilde{A}, \tilde{P}, \tilde{R}, \gamma, O)\) for the  LLM-SOARL framework and divide it into three parts:

\begin{itemize}
    \item  \(C\) denotes the set of natural language constraints (e.g., "Do not pass through plants and printers"), serving as high-level behavioral guidelines to ensure the agent's compliance and providing a semantic interaction foundation for the framework.
    \item We define a planning problem with symbolic actions \((I, G, P, A, S, F, O)\): \(I\) is the initial symbolic state, and \(G\) is the dynamic goal state that drives the agent to pursue higher cumulative rewards. \(P\) is a set of propositions with prior knowledge, represented by the Planning Domain Definition Language (PDDL), and is used to describe the symbolic state space \(S\) (where \(S \subseteq 2^P\)). \(A\) is a set of action models that map one symbolic state to another (\(S \times A \to S\)), and each model is automatically learned by the meta-controller through symbolic state pairs. \(F\) is a state mapping function (\(F: \tilde{s} \to 2^P\)) responsible for converting high-dimensional primitive states \(\tilde{s}\) into interpretable symbolic states \(s\). \(O\) is the option set, including symbolic options (encapsulating preconditions, low-level policies, and execution effects) and a global exploration option, providing support for hierarchical learning and the discovery of new action models.
    
    \item We define the underlying decision-making problem through a Markov Decision Process (MDP) \((\tilde{S}, \tilde{A}, \tilde{P}, \tilde{R}, \gamma)\). We denote symbolic actions and states as \(a\) and \(s\) respectively, while primitive actions and states as \(\tilde{a}\) and \(\tilde{s}\). Here, \(\tilde{a}\) and \(\tilde{s}\) are both obtained from direct interaction with the environment. \(\tilde{P}\) describes the transition probability between primitive states, \(\tilde{R}\) provides immediate rewards for each primitive state-action combination, and \(\gamma\) is a discount factor that balances the importance of immediate and future rewards. 
\end{itemize}
The framework aims to autonomously learn interpretable action models \(A\) and options \(O\), which can be utilized to generate option sequences. It achieves cross-task skill reuse through LLM-driven semantic processing, ensures constraint compliance via real-time intervention, and ultimately maximizes the cumulative reward.

\section{Preliminaries}
This section introduces the fundamental concepts used in this paper, including symbolic planning with PDDL, reinforcement learning, the option framework, and LLMs.
\subsection{Symbolic Planning with PDDL}

In the Planning Domain Definition Language (PDDL), states are represented as proposition sets, termed symbolic states herein to distinguish them from reinforcement learning (RL) states. Propositions describe world properties: for a symbolic state \(s\), \(p \in s\) if proposition \(p\) holds, otherwise \(not\ p \in s\). An action model is defined as a tuple(\(name, pre^+, pre^-, eff^+, eff^-)\), where \(name\) is the action identifier, \((pre^+, pre^-)\) are positive/negative preconditions, and \((eff^+, eff^-)\) are positive/negative effects.
Action \(a\) is executable if \(pre^+ \subset s\) and \(s \cap pre^- = \emptyset\). The next state after execution is $s' = ((s - eff^-) \cup eff^+)$. 
A planning domain \(D = (P, A)\) includes proposition set \(P\) (state space) and action set \(A\) (action space). A tuple \((s, a, s')\) denotes a symbolic transition. A planning problem is \((I, P, A, G)\) (initial state \(I\), goal state \(G\)), and its solution is a sequential plan \(\pi\) generating a transition trace from \(I\) to \(G\). To maximize cumulative reward, we use the Metric-FF planner (Hoffmann 2002) for continuous metric tasks.
\subsection{Reinforcement Learning}
A Markov Decision Process (MDP) is defined as a tuple \((\tilde{S}, \tilde{A}, P_{\tilde{s}\tilde{s}'}^{\tilde{a}}, r_{\tilde{s}}^{\tilde{a}}, \gamma)\), where \(\tilde{S}\) and \(\tilde{A}\) denote the state and action spaces, respectively; \(P_{\tilde{s}\tilde{s}'}^{\tilde{a}}\) is the transition probability from state \(\tilde{s} \in \tilde{S}\) to \(\tilde{s}' \in \tilde{S}\) via action \(\tilde{a} \in \tilde{A}\); \(r_{\tilde{s}}^{\tilde{a}}\) is the immediate reward of taking \(\tilde{a}\) at \(\tilde{s}\); and \(\gamma \in [0,1)\) is a discount factor.

The goal of RL is to learn a policy \(\pi: \tilde{S} \to \tilde{A}\) that maximizes the expected return, computed by:
\begin{equation}
  V_{\pi}(\tilde{s}) = \mathbb{E}_{\pi}\left[\sum_{t=0}^{\infty} \gamma^t r_t \mid \tilde{s}_0 = \tilde{s}\right]  
\end{equation}
where \(r_t\) is the reward at time step \(t\) when following \(\pi\) from initial state \(\tilde{s}_0 = \tilde{s}\).

The state-action value function is defined as:
\begin{equation}
  Q_{\pi}(\tilde{s}, \tilde{a}) = \mathbb{E}_{\pi}\left[\sum_{t=0}^{\infty} \gamma^t r_t \mid \tilde{s}_0 = \tilde{s}, \tilde{a}_0 = \tilde{a}\right]  
\end{equation}
representing the expected return of taking action \(\tilde{a}\) at state \(\tilde{s}\) and following \(\pi\) thereafter.
\subsection{Option Framework}
Hierarchical Reinforcement Learning (HRL) extends traditional RL by introducing temporally abstract macro-actions, which are formally modeled as \textit{options} in the option framework. An option \(o\) is defined as a triple \((I_o(\tilde{s}), \pi_o(\tilde{s}), \beta_o(\tilde{s}))\), where:
The initiation condition \(I_o(\tilde{s})\) determines whether \(o\) can be executed at state \(\tilde{s}\);The termination condition \(\beta_o(\tilde{s})\) decides if the execution of \(o\) terminates at state \(\tilde{s}\); The policy \(\pi_o(\tilde{s})\) maps state \(\tilde{s}\) to low-level primitive actions.

In this framework, the learning process is divided into two levels: the high-level meta-controller learns to select optimal options, while the low-level controller optimizes policies to execute the selected options. 

\subsection{Large Language Models}
LLMs are deep neural networks trained on large-scale text corpora and exhibit strong capabilities in semantic understanding, abstraction, and reasoning. Recently, LLMs have been explored for semantic parsing, high-level planning \cite{zhang2023planning}, and constraint reasoning, serving as semantic interfaces in reinforcement learning and symbolic systems. Rather than replacing learning-based controllers, LLMs are better viewed as complementary modules that enhance interpretability and generalization.
\begin{figure*}[htbp]
    \centering
    \includegraphics[width=1\textwidth]{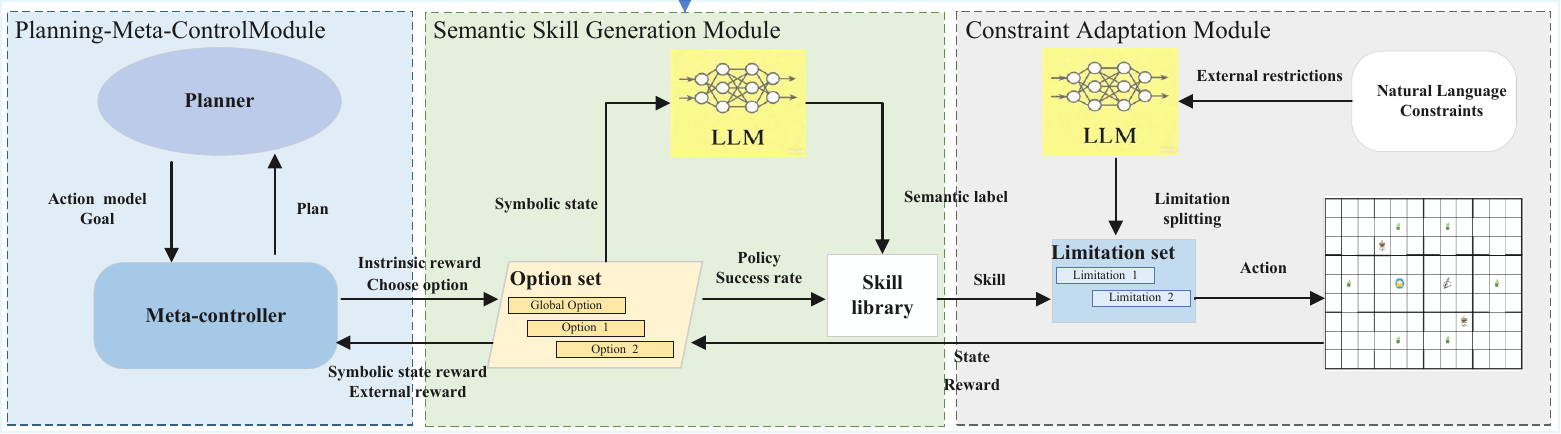}
    \caption{The  LLM-SOARL framework}
    \label{fig: LLM-SOARL}
\end{figure*}

\section{The  LLM-SOARL Framework}
As shown in Figure \ref{fig: LLM-SOARL}, the  LLM-SOARL framework is a closed-loop system, consisting of three core modules:
\begin{enumerate}
    \item \textbf{Planning-Meta-Control Module} integrates a symbolic planner and a meta-controller, responsible for generating reward-optimized plans, learning action models, formulating dynamic goals, and selecting optimal options .
    \item \textbf{Semantic Skill Module} is powered by LLMs, integrating with an option set. Through semantic annotation, retrieval, and transfer of option policies, it achieves cross-task skill reuse while undertaking the execution function of interacting with the environment.
    \item \textbf{Constraint adaptation Module}: A constraint enforcement module driven by LLMs, which parses natural language constraints into executable rules and conducts real-time monitoring to ensure behavioral compliance.
\end{enumerate}

The framework operates through continuous iterative loops: it receives user inputs such as natural language constraints and task goals, relies on the three core modules to complete planning, learning, skill reuse, and constraint enforcement, and continuously optimizes models and policies through environmental interaction feedback. Ultimately, it achieves efficient, compliant, and interpretable decision-making in structured environments.

\subsection{Planning-Meta-Control Module}
The Planning-Meta-Control module \cite{jin2022creativity} encompasses the autonomous learning of action models, the formulation of dynamic planning goals, the scheduling of optimal options, and the provision of intrinsic reward signals to the low-level policy. In the remainder of this sub-section, we briefly introduce the symbolic Planning-Meta-Control framework proposed in \cite{jin2022creativity}. 

Specifically, the framework first initializes an option set $O$ only including a global option $o_G$. 
A symbolic option $so$ is formally defined as a tuple $(pre, \pi, eff)$, where $\pi$ represents the low-level execution policy, and $pre$ and $eff$ denote the symbolic preconditions and execution effects, respectively. Given a symbolic option $so$ and a high-dimension state $\tilde{s}$, the module computes initial conditions $I_{so}(\tilde{s})$ and the termination condition $\beta_{so}(\tilde{s})$ according to effects and preconditions. The global option $o_G$ is assumed to be applicable to all states (i.e., $I_G(\tilde{s}) = True$) and employs a random policy $\pi = \text{random}(\tilde{A})$ to explore the environment until a change in the symbolic state occurs. 

When an episode $t$ begins, we first get an initial state $\tilde{s}_0$ from environments, compute the symbolic initial state $I$ by mapping function $F$, and record the best plan. Then meta-controller updates action models $A$, symbolic options set $O$, their mapping function $F_{A,O}$ transferring action models to options, and the planning goal $G$. According to symbolic state pairs $(s_{t}, s_{t+1})$ as well as their rewards generated during interaction, the module learns action $a_i$ according to positive/negative preconditions and effects. 

Given action models $A$ and planning goal $G$, a planner (Metric-FF in this paper) generates a new plan $\Pi_t$ whose quality is higher than $\Pi_{t-1}$. As for each action model $a_i$ in plan $\Pi_t$, the meta-controller chooses a corresponding symbolic option $o_j$ by $F_{A,O}$. Then the controller interacts with environment by performing Deep Q-Learning, executes the action chosen by $o_j$'s inner policy and stores experience into $o_j$'s replay buffer until $o_j$ terminates. 

As for each action model $a_i$, the meta-controller selects a symbolic option from the option set by $o_j=F_{A,O}(a_i)$ and gets a series of options $(o_0,o_1,\dots,o_n)$. If all action models in $\Pi_t$ successfully finish, which indicates the chosen symbolic options are executed sequentially, and termination conditions are satisfied. Then the meta-controller would choose the global option $o_G$ to explore the environment. For each option $o_i = (pre_i,\pi_i,\beta_i)$, the intrinsic rewards is defined by:
\begin{equation}
r_i(\tilde{s}) = \begin{cases} 
r + \psi & \text{if } \beta_i(\tilde{s}') = True \\ 
r & \text{otherwise} 
\end{cases}
\end{equation}

where $\psi$ is a predefined constant serving as a terminal reward for successful option execution. 

After that, we get $o_j$'s initial symbolic state $s_{1}$, a terminal symbolic state $s_{2}$, and an extrinsic reward $r_e$. the Meta-controller assigns a reward weight $\rho_i$ to each action model $a_i$. This weight integrates the historical average external reward and an exploration reward $r_e$:
\begin{equation}
\rho_i = \text{mean}(R[(s_{t}, s_{t+1})_i]) + r_e
\end{equation}
The exploration reward $r_e$ aims at encouraging the selection of action models with lower success rates ($sr$), calculated by:
\begin{equation}
r_e = \begin{cases} 
c(1 - sr[i]) & \text{if } a_i \text{ is under exploration} \\ 
0 & \text{otherwise} 
\end{cases}
\end{equation}
where $c$ is a predefined constant.

In this way, we compute symbolic state pairs and their extrinsic rewards one by one and record these mappings in a dictionary $R$. Finally, quality $q$ of plan $\Pi_t$ is defined as the accumulated sum of extrinsic rewards. 

If the environment isn't finished after executing $\Pi_t$, the meta-controller chooses the global option $o_G$ to explore new symbolic state pairs in the environment. $o_G$ stops exploring when the computed symbolic state changes, and we calculate a symbolic state pair $(s_{1},s_{2})$ and its external reward $r_e$. If $(s_{1},s_{2})$ is a new symbolic state pair, we add it to $R$. This process repeats until the environment is terminated. Finally, when an episode ends, we train options in $O$ and calculate the success ratio for each action model.

\subsection{Semantic Skill Generation Module}

This module aims to learn policy $\pi_o$ considering semantics to construct a skill reuse system, similar to recent trends in using LLMs for embodied control, which allows agents to use the skill when facing similar events. Specifically, the Semantic Skill Generation Module first explores options according to the Option Set and the current state. Then, we transmit the state changes encoding to LLMs based on hand-made prompts. LLMs then generate structured Semantic Labels $l_o$ for the current policy $\pi_o$ accordingly, which include the information of the initial and target objects. Subsequently, the module filters effective skills, based on the policy's success rate $sr(o)$, given by interaction with environments, and stores them in the Skill Library. When utilizing Skill Library, we compare the policy and the one with the same semantic labels stored in the library based on the accuracy, and choose the more precise one as the current policy. 


\subsubsection{Semantic Labels Generation}

\begin{figure}[htbp]
    \centering
    \includegraphics[width=0.45\textwidth]{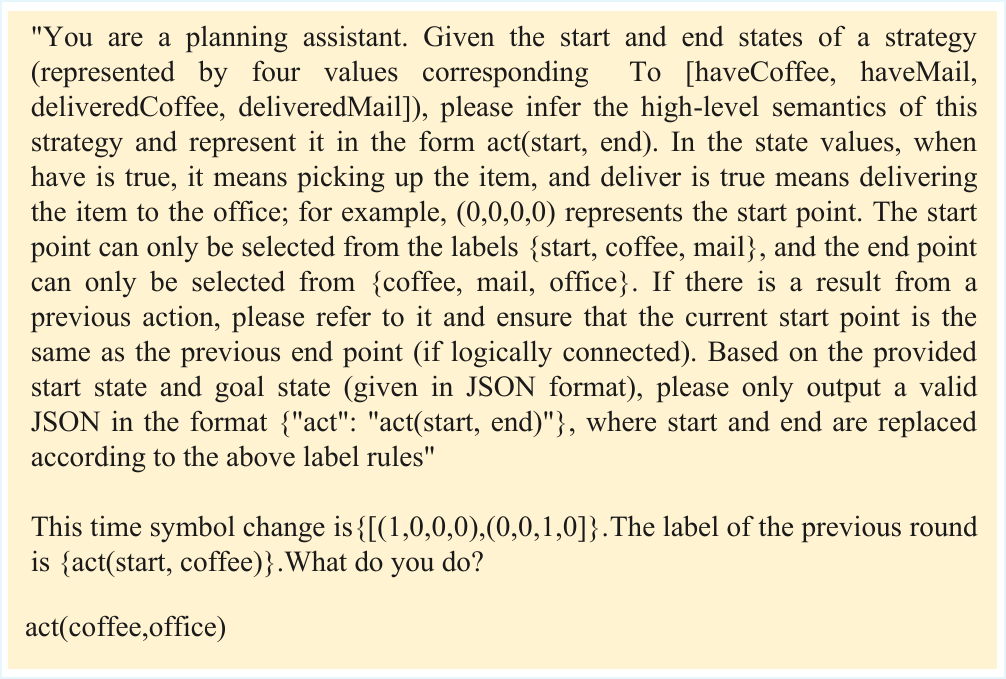}
    \caption{Generate semantic labels using LLMs}
    \label{fig:Semantic-Label}
\end{figure}

To generate semantic labels, we generate options and compute state traces for LLMs. To extract semantic information from raw data, the module first explores an option $o_i$ based on the current state $s$ and next state $s'$ updated by $o_i$. Then we utilize LLM to abstract these symbolic state changes into structured Semantic labels ($l_{o_i}$), where the prompt template includes domain information, the generation requirements, the encoding of current state $s$ and goal state $s'$, and the last semantic label. To ensure logical consistency and interpretability, $l_{o_i}$ adopts a ``Predicate-Argument'' structure.  
\emph{As illustrated in Figure \ref{fig:Semantic-Label} within the Office World domain, the symbolic state is represented by a four-dimensional vector $[\textit{haveCoffee}, \textit{haveMail}, \textit{deliveredCoffee}, \textit{deliveredMail}]$. Upon the completion of an option, if the symbolic state transitions from $s=(1,0,0,0)$ (indicating possession of coffee) to $s'=(0,0,1,0)$ (indicating the coffee has been delivered), the $M_{LLM}$ captures this state differential. By integrating contextual logic, the model abstracts this low-level numerical transition into a structured semantic label $l_{sem}=\text{Act}(\textit{coffee}, \textit{office})$. This instance demonstrates how the module transforms implicit state interactions into explicit, interpretable instructions adhering to the ``Predicate-Argument'' structure.}


\subsubsection{ Skill Library Expansion}

After obtaining semantic labels $l_{o_{i}}$, we check whether it is effective. 
We compute the success rate $sr(o_i)$ of the exploration policy $\pi_o{_i}(\tilde{s})$ by interacting with environments. If the success rate exceeds a preset confidence threshold $\tau$, we regard it as an effective policy and encapsulate the generated semantic label with the option $o_i$ to update the Skill library ($L_{skill}$), establishing an explicit semantic index for each low-level executable option:
\begin{equation}
L_{skill} \leftarrow L_{skill} \cup \{(\pi_{o_i}(\tilde{s}), l_{o_{i}})    
\end{equation}

\subsubsection{Skill Reusage}
To reuse stored skills to avoid repeatedly learning during the exploration phase, when the agent discovers a new symbolic option $o_i$, it first uses the LLMs to generate the corresponding semantic label $l_{o_i}$ and then queries $L_{skill}$ for a matching policy. If a semantically equivalent label exists in $L_{skill}$, the agent retrieves the corresponding option policy $\pi_{o_i}$. Otherwise, the system regards $o_i$ as a newfound
option and stores it in the library if it meets the condition that the success rate of policy  $\pi_{o_i}$ reaches the threshold $\tau$. 


\subsection{Constraint Adaptation Module}

This module is designed to formulate adaptive rules based on intuitive linguistic instructions. Unlike traditional constrained policy optimization.  As shown in the Algorithm \ref{alg:constraint_adaptation}, our workflow operates by transmitting natural language as constraints to LLMs to extract abstract entities via handcrafted prompt templates, which are subsequently transformed into a set of corresponding atomic propositions, i.e., the limitation set. This set is then injected into a Reward Machine to monitor state feedback in real-time and trigger penalty signals for violations. 


\begin{algorithm}[tb]
    \caption{LLM-driven Constraint Adaptation Algorithm}
    \label{alg:constraint_adaptation}
    \begin{algorithmic}[1] 
        \REQUIRE Natural language constraint $C$, LLM, Mapping function $D$, Labeling function $L$
        \ENSURE Policy $\pi$ compliant with constraints
        
        \STATE \textbf{Initialization}: 
        \STATE Extract abstract forbidden entities via LLM:
        \STATE $E_{b} \leftarrow \text{Extract}(LLM(C))$
        \STATE Construct Limitation Set via limitation splitting:
        \STATE $P_{b} \leftarrow D(E_{b})$
        \STATE Initialize Reward Machine state $u_0 \leftarrow u_{init}$
        
        \STATE \textbf{Execution Loop}:
        \FOR{each time step $t=0, 1, \dots$}
            \STATE Agent executes action $a_t$, observes $s_{t+1}$ and intrinsic reward $r_{env}$
            \STATE Obtain atomic propositions of the current state:
            \STATE $P_{current} \leftarrow L(s_{t+1})$
            
            \STATE Calculate violation indicator :
            \STATE $I_{viol} \leftarrow \mathbb{I}(P_{current} \cap P_{b} \neq \emptyset)$
            
            \STATE Update Reward Machine state :
            \IF{$I_{viol} = 1$}
                \STATE $u_{t+1} \leftarrow u_{broken}$ \COMMENT{Violation detected}
                \STATE $r_t \leftarrow r_{penalty}$
                \STATE \textbf{break} current episode
            \ELSE
                \STATE $u_{t+1} \leftarrow u_t$ \COMMENT{Safe transition}
                \STATE $r_t \leftarrow r_{env}$
            \ENDIF
            
            \STATE Update policy $\pi$ using tuple $(s_t, a_t, r_t, s_{t+1})$
        \ENDFOR
    \end{algorithmic}
\end{algorithm}

\subsubsection{LLM and External Restrictions}
To address these External restrictions, the module invokes the LLM assistant to perform semantic parsing. The LLM utilizes its natural language understanding capabilities to extract specific abstract entities $E_{b}$ from the unstructured instructions (The corresponding prompt is illustrated in the supplementary materials). Taking the example in Figure \ref{fig:Natural-Language}, the extracted entities of ``We need to be careful not to bump into any plants and printers'' are \{\text{printer}, \text{plant}\}.  
By extracting these entities, the model effectively filters out irrelevant information, establishing a preliminary mapping from high-level natural language semantics to concrete environmental conceptual entities.

\subsubsection{Limitation splitting and Limitation set}
After obtaining the abstract entities, the system executes the Limitation splitting operation to disassemble complex linguistic descriptions into atomic propositions recognizable by the environment. The module obtains a specific set of atomic propositions $P_{b} \subseteq P$, the limitation set, including all propositions related to entities $E_{b}$. This process concretizes abstract linguistic constraints into logical conditions that reflect environmental states, allowing agents to obtain feedback from the Reward Machine. These propositions can be formally injected into the state transition function as criteria for identifying violation states. 

During exploration, we use the limitation set to assist in predicting rewards before executing actions according to explored experience. 

\subsubsection{Action, State and Reward}
During the exploration phase of the agent, the system implements real-time constraint verification by monitoring the feedback following the interaction between the Action and the environment. When the agent executes Action $\tilde{a}$ at time $t$ and transitions to the next state $\tilde{s}_{t+1}$, the environment returns the set of atomic propositions satisfied by the current State. The system computes the intersection of the current state and the Limitation set ($P_{b}$), computed by:

\begin{equation}
   I_{viol}(\tilde{s}_{t+1}) = F(\tilde{s}_{t+1}) \cap P_{b} 
\end{equation}
where $F(\tilde{s}_{t+1})$ maps the $\tilde{s}_{t+1}$ to the high-level representations with atomic propositions. If $F(\tilde{s}_{t+1})$ is empty, the limitation set is inapplicable to the state, and we continue exploring and interacting with environments. Otherwise, we consider that historical experience could help predict future exploration. We compute a predicted reward $r_p$ or punishment as assistance according to the experienced reward. 
\begin{equation}
    r_p = \lambda\tilde{R}(\tilde{s},\tilde{a})
\end{equation}
where $\lambda$ is a hyperparameter, which is $0.1$ in this paper. 


This state transition triggers a corresponding Reward penalty or termination signal, thereby ensuring that constraints are strictly enforced and fed back at the task execution level. The rewards or penalties derived from the Reward Machine are utilized to update the policy $\pi$, thereby enabling it to adapt more effectively to the task requirements.

\section{Experiment}
We evaluate our approach on two domains, Office World and Montezuma’s Revenge, in terms of data efficiency, constraint compliance, and cross-task transfer.
\begin{figure*}[htbp]
     \centering
     \includegraphics[width=0.9\textwidth]{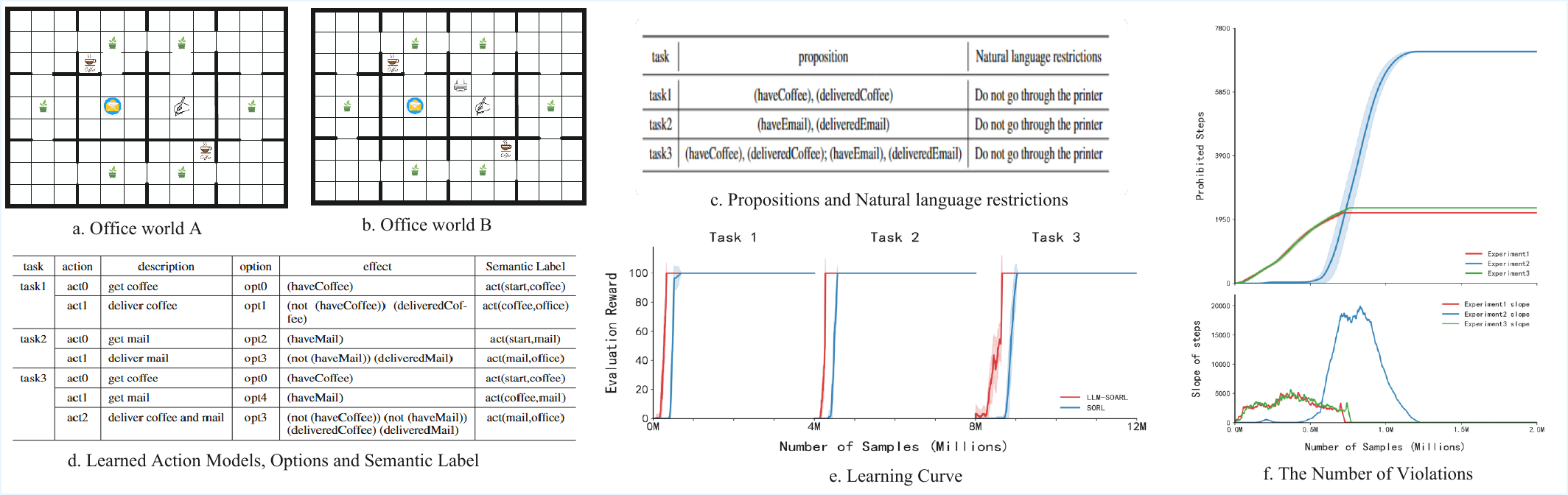}
    \caption{The Office World Result}
    \label{fig: office world result}
\end{figure*}

\subsection{Office World}
To validate the efficacy of our method, we employ the "Office World"\cite{toro2018advice} domain to evaluate the LLM-SOARL. In this experimental setting, the agent is initialized at a randomly generated coordinate, with an action space comprising movements in the four cardinal directions; however, the trajectory is physically constrained by wall obstacles. The interaction logic within the environment is as follows: grid cells distinguished by a black cup icon represent coffee resources, whereas those marked with a blue envelope denote mail. Reaching these specific regions triggers a pick-up operation. Furthermore, the region marked with a "hand holding a pen" icon is designated as the office; the agent is required to navigate to this location to execute the delivery tasks for coffee or mail.
\subsubsection{Setup}
In our experiments, the agent's initial position is randomized at the beginning of each episode to ensure robust evaluation. To evaluate the adaptive policy, we construct two scenes in the office world domain, the office world A and the office world B, as shown in Fig. 4(a) and (b). The office world A contains coffee and mail to be delivered and plants to be avoided. The office world B extra includes a printer to be avoided, where the other settings are the same. 

We defined three distinct tasks to assess performance: as shown in Fig.4(c), Task 1 and Task 2 require the agent to deliver a cup of coffee or a piece of mail to the office, respectively, while Task 3 presents a more complex composite goal requiring the delivery of both objects. We benchmarked our proposed LLM-SOARL framework against SORL\cite{jin2022creativity}. Given the finite nature of the state and action spaces in this domain, both the high-level meta-controller and low-level controllers for these approaches were implemented using tabular Q-learning (Q-tables) to ensure a fair comparison.
\subsubsection{Results}
We evaluate our approaches in terms of data efficiency, constraint compliance, and cross-task transferability.
\begin{itemize}
    \item \textbf{Data Efficiency}  To evaluate data efficiency, we conducted experiments under both progressive (Task1/2$\rightarrow$ Task3) and reverse-inclusion (Task3 $\rightarrow$ Task1/2) settings in the office world B. As shown in Fig.4(e), LLM-SOARL achieves significantly faster convergence than SORL in both scenarios. Notably, the rapid performance jump observed immediately after task switching empirically proves the effectiveness of reusing prior knowledge to reduce the samples required for learning new tasks.
    
    \item \textbf{Policy Adaptation for Extra Constraints}  To verify the effectiveness of our Constraint Adaptation Module in this modified environment, we designed three experimental settings: (1) Sequential Transfer in the office world B, where it learns Task1 and Task2 before Task3 (Experiment 1 in the figure); (2) Learning from Scratch in the office world B, where the agent learns Task3 directly (Experiment 2 in the figure); and (3) Learning based on previous experience, utilizing the agent have been trained in the office world A to train in the office world B. As shown in Fig.4(f), the cumulative violations in three cases eventually plateau, indicating that the agent successfully internalizes the natural language constraints. Moreover, the Sequential Transfer method results in significantly fewer total violations compared to learning from scratch, proving that reusing pre-trained skills effectively minimizes unsafe exploration. 
    
    \item \textbf{Cross-Task Transferability Mechanism}  The superior data efficiency stems from the Semantic Label Matching mechanism illustrated in Fig.4(d). By leveraging LLMs to generate interpretable labels (e.g., $act(start, coffee)$), the agent identifies that the "get coffee" sub-goal in the new task is semantically identical to a pre-trained option from the Skill Library. This precise semantic alignment enables the agent to directly retrieve and execute the optimal policy without redundant exploration, ensuring that learned skills are task-agnostic and transferable.
\end{itemize}  
\begin{figure}[htbp]
     \centering
     \includegraphics[width=0.45\textwidth]{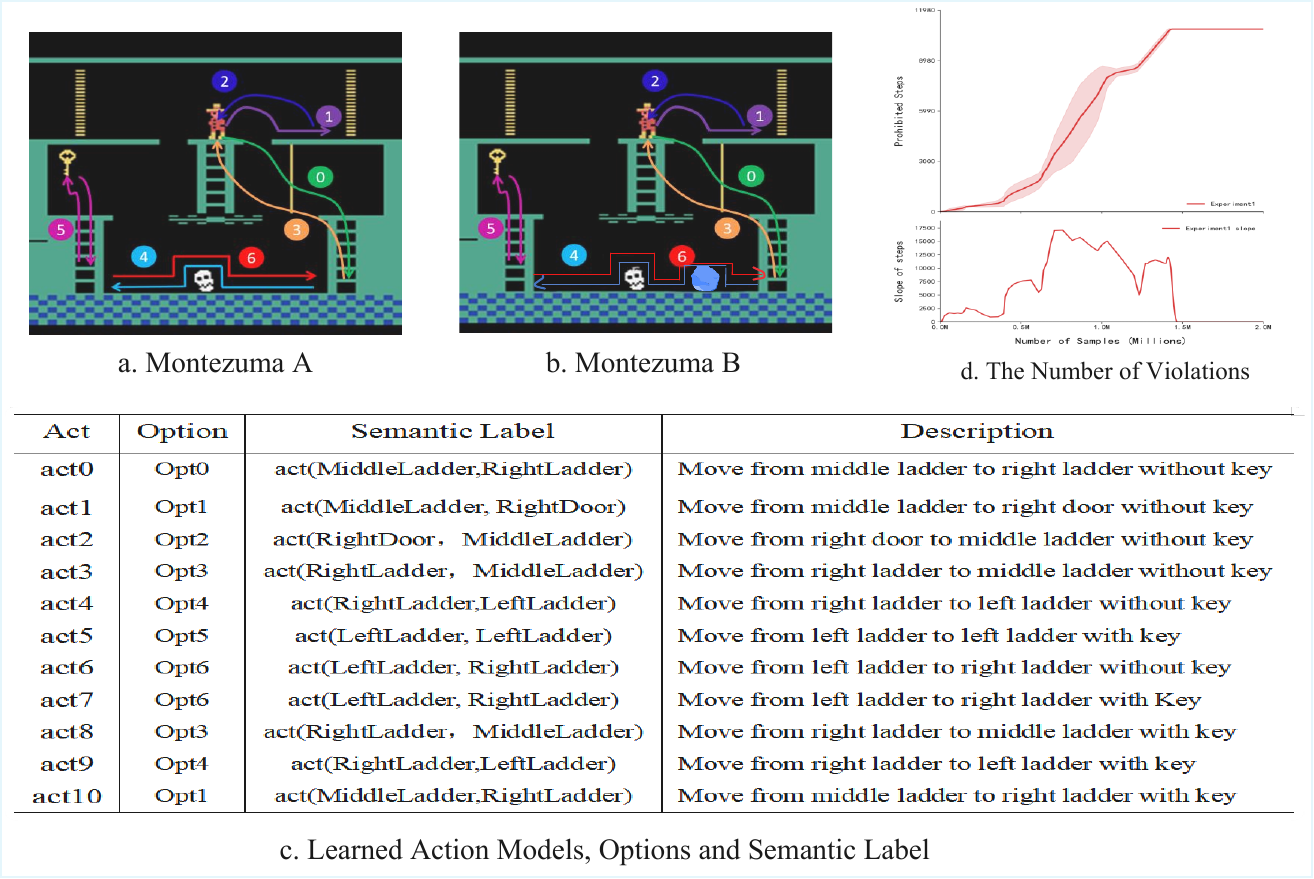}
    \caption{The Montezuma's Revenge Result}
    \label{fig: Montezuma result}
\end{figure}
\subsection{Montezuma's Revenge}
To further validate the efficacy of our method in high-dimensional environments characterized by sparse and delayed rewards, we employ Montezuma’s Revenge as a challenging evaluation platform. In this experimental setting, the agent is tasked with navigating the first room of the game. The interaction logic defines strict reward conditions: the agent receives positive feedback only upon fetching the key (+100) or opening the door (+300), with the optimal trajectory yielding a maximum cumulative reward of +400.
\subsubsection{Setup}
Our framework is implemented within a standardized option-based HRL architecture. For the low-level controller, we adopted the neural network architecture used in \cite{kulkarni2016hierarchical} and trained it utilizing Double Q-learning \cite{van2016deep} and Prioritized Experience Replay \cite{schaul2015prioritized} to enhance sample efficiency. LLM-SOARL utilize a planner for high-level policy generation.
Regarding the hyperparameter configuration, the intrinsic reward constant $\phi$ was set to 100. We established a maximum episode length of 500 steps and set the success rate threshold for option validation at 0.95. To construct the abstract state space, we defined five atomic propositions: four representing the agent's location (\textit{MiddleLadder}, \textit{RightDoor}, \textit{LeftLadder}, \textit{RightLadder}) and one indicating the object status (\textit{Key}).
\subsubsection{Results}
Having validated the cross-task transferability in the Office World domain, we focus here on evaluating the performance of LLM-SOARL in terms of intra-task skill reuse (facilitated by semantic skill discovery) and constraint compliance.
\begin{itemize}
 \item \textbf{Intra-task Skill Reuse} The experimental results demonstrate the framework's capability to discover and reuse skills within a single complex task. As illustrated in Fig.5(c), the LLM-driven module generates interpretable semantic labels for learned options using a "Predicate-Argument" structure. A critical observation is the semantic equivalence between different options. For instance, both $act3$ (executed without a key) and $act8$ (executed with a key) are annotated with the same label: act(RightLadder, MiddleLadder). This indicates that the LLM successfully abstracted the navigation policy from the low-level state variable (holding a key), allowing the agent to reuse the same "Move from right ladder to middle ladder" skill across different task stages. This mechanism significantly reduces redundant learning of identical physical movements.
 
\item \textbf{Constraint Compliance} To evaluate the Constraint Adaptation Module, we modified the standard environment (Fig.5(a)) by introducing a blue stone obstacle (Fig.5(b)) and imposing a corresponding natural language constraint: ''Do not touch the stone. Fig.5(d) visualizes the agent's adaptation process to the natural language constraint. The "Prohibited Steps" curve shows the cumulative number of violations over time. In the initial 0.7M samples, the violation rate increases as the agent interacts with the environment. However, as the Constraint Adaptation Module translates the linguistic instruction into penalty signals within the Reward Machine, the "Slope of steps" (representing the rate of new violations) peaks and then sharply declines. By 1.5M samples, the slope drops to zero and the cumulative violations plateau, confirming that the agent has successfully internalized the constraint and achieved zero-violation performance while completing the task.
\end{itemize}  

\section{Conclusions}
This paper proposes LLM-SOARL, a novel closed-loop framework integrating LLMs, symbolic planning, and DRL. The framework achieves automatic discovery and reuse of general skills through a Semantic Skill Generation module and transforms unstructured natural language instructions into executable real-time safety monitoring rules via a Constraint Adaptation module. Experimental results demonstrate that LLM-SOARL significantly outperforms existing baseline methods in terms of sample efficiency, constraint compliance, and cross-task transferability, providing an efficient and safe solution for intelligent decision-making in scenarios with sparse rewards and complex semantic interactions.
\bibliographystyle{named}

\bibliography{ijcai26}

@article{andrychowicz2020learning,
  title={Learning dexterous in-hand manipulation},
  author={Andrychowicz, OpenAI: Marcin and Baker, Bowen and Chociej, Maciek and Jozefowicz, Rafal and McGrew, Bob and Pachocki, Jakub and Petron, Arthur and Plappert, Matthias and Powell, Glenn and Ray, Alex and others},
  journal={The International Journal of Robotics Research},
  volume={39},
  number={1},
  pages={3--20},
  year={2020},
  publisher={SAGE Publications Sage UK: London, England}
}

@article{aradi2020survey,
  title={Survey of deep reinforcement learning for motion planning of autonomous vehicles},
  author={Aradi, Szil{\'a}rd},
  journal={IEEE Transactions on Intelligent Transportation Systems},
  volume={23},
  number={2},
  pages={740--759},
  year={2020},
  publisher={Ieee}
}

@article{degrave2022magnetic,
  title={Magnetic control of tokamak plasmas through deep reinforcement learning},
  author={Degrave, Jonas and Felici, Federico and Buchli, Jonas and Neunert, Michael and Tracey, Brendan and Carpanese, Francesco and Ewalds, Timo and Hafner, Roland and Abdolmaleki, Abbas and de Las Casas, Diego and others},
  journal={Nature},
  volume={602},
  number={7897},
  pages={414--419},
  year={2022},
  publisher={Nature Publishing Group UK London}
}

@inproceedings{du2023guiding,
  title={Guiding pretraining in reinforcement learning with large language models},
  author={Du, Yuqing and Watkins, Olivia and Wang, Zihan and Colas, C{\'e}dric and Darrell, Trevor and Abbeel, Pieter and Gupta, Abhishek and Andreas, Jacob},
  booktitle={International Conference on Machine Learning},
  pages={8657--8677},
  year={2023},
  organization={PMLR}
}

@article{fawzi2022discovering,
  title={Discovering faster matrix multiplication algorithms with reinforcement learning},
  author={Fawzi, Alhussein and Balog, Matej and Huang, Aja and Hubert, Thomas and Romera-Paredes, Bernardino and Barekatain, Mohammadamin and Novikov, Alexander and R. Ruiz, Francisco J and Schrittwieser, Julian and Swirszcz, Grzegorz and others},
  journal={Nature},
  volume={610},
  number={7930},
  pages={47--53},
  year={2022},
  publisher={Nature Publishing Group UK London}
}

@inproceedings{illanes2020symbolic,
  title={Symbolic plans as high-level instructions for reinforcement learning},
  author={Illanes, Le{\'o}n and Yan, Xi and Icarte, Rodrigo Toro and McIlraith, Sheila A},
  booktitle={Proceedings of the international conference on automated planning and scheduling},
  volume={30},
  pages={540--550},
  year={2020}
}

@inproceedings{jin2022creativity,
  title={Creativity of ai: Automatic symbolic option discovery for facilitating deep reinforcement learning},
  author={Jin, Mu and Ma, Zhihao and Jin, Kebing and Zhuo, Hankz Hankui and Chen, Chen and Yu, Chao},
  booktitle={Proceedings of the AAAI conference on artificial intelligence},
  volume={36},
  number={6},
  pages={7042--7050},
  year={2022}
}

@article{kulkarni2016hierarchical,
  title={Hierarchical deep reinforcement learning: Integrating temporal abstraction and intrinsic motivation},
  author={Kulkarni, Tejas D and Narasimhan, Karthik and Saeedi, Ardavan and Tenenbaum, Josh},
  journal={Advances in neural information processing systems},
  volume={29},
  year={2016}
}

@article{mirhoseini2021graph,
  title={A graph placement methodology for fast chip design},
  author={Mirhoseini, Azalia and Goldie, Anna and Yazgan, Mustafa and Jiang, Joe Wenjie and Songhori, Ebrahim and Wang, Shen and Lee, Young-Joon and Johnson, Eric and Pathak, Omkar and Nova, Azade and others},
  journal={Nature},
  volume={594},
  number={7862},
  pages={207--212},
  year={2021},
  publisher={Nature Publishing Group UK London}
}

@article{mnih2015human,
  title={Human-level control through deep reinforcement learning},
  author={Mnih, Volodymyr and Kavukcuoglu, Koray and Silver, David and Rusu, Andrei A and Veness, Joel and Bellemare, Marc G and Graves, Alex and Riedmiller, Martin and Fidjeland, Andreas K and Ostrovski, Georg and others},
  journal={nature},
  volume={518},
  number={7540},
  pages={529--533},
  year={2015},
  publisher={Nature Publishing Group}
}

@article{schaul2015prioritized,
  title={Prioritized experience replay},
  author={Schaul, Tom and Quan, John and Antonoglou, Ioannis and Silver, David},
  journal={arXiv preprint arXiv:1511.05952},
  year={2015}
}

@article{shinn2023reflexion,
  title={Reflexion: Language agents with verbal reinforcement learning},
  author={Shinn, Noah and Cassano, Federico and Gopinath, Ashwin and Narasimhan, Karthik and Yao, Shunyu},
  journal={Advances in neural information processing systems},
  volume={36},
  pages={8634--8652},
  year={2023}
}

@article{tam2022semantic,
  title={Semantic exploration from language abstractions and pretrained representations},
  author={Tam, Allison and Rabinowitz, Neil and Lampinen, Andrew and Roy, Nicholas A and Chan, Stephanie and Strouse, DJ and Wang, Jane and Banino, Andrea and Hill, Felix},
  journal={Advances in neural information processing systems},
  volume={35},
  pages={25377--25389},
  year={2022}
}

@inproceedings{toro2018advice,
  title={Advice-based exploration in model-based reinforcement learning},
  author={Toro Icarte, Rodrigo and Klassen, Toryn Q and Valenzano, Richard Anthony and McIlraith, Sheila A},
  booktitle={Canadian Conference on Artificial Intelligence},
  pages={72--83},
  year={2018},
  organization={Springer}
}

@inproceedings{van2016deep,
  title={Deep reinforcement learning with double q-learning},
  author={Van Hasselt, Hado and Guez, Arthur and Silver, David},
  booktitle={Proceedings of the AAAI conference on artificial intelligence},
  volume={30},
  number={1},
  year={2016}
}

@article{xi2025rise,
  title={The rise and potential of large language model based agents: A survey},
  author={Xi, Zhiheng and Chen, Wenxiang and Guo, Xin and He, Wei and Ding, Yiwen and Hong, Boyang and Zhang, Ming and Wang, Junzhe and Jin, Senjie and Zhou, Enyu and others},
  journal={Science China Information Sciences},
  volume={68},
  number={2},
  pages={121101},
  year={2025},
  publisher={Springer}
}

@article{zhang2023planning,
  title={Planning with logical graph-based language model for instruction generation},
  author={Zhang, Fan and Jin, Kebing and Zhuo, Hankz Hankui},
  journal={arXiv preprint arXiv:2308.13782},
  year={2023}
}

\end{document}